%% file: aaai2023_conference.tex
\documentclass[letterpaper]{article} 
\usepackage{aaai23}  
\usepackage{times}  
\usepackage{helvet}  
\usepackage{courier}  
\usepackage[hyphens]{url}  
\usepackage{graphicx} 
\usepackage{natbib}  
\usepackage{caption} 
\usepackage{algorithm}
\usepackage{algorithmic}

\usepackage{newfloat}
\usepackage{listings}
\DeclareCaptionStyle{ruled}{labelfont=normalfont,labelsep=colon,strut=off} 
\floatstyle{ruled}
\newfloat{listing}{tb}{lst}{}
\floatname{listing}{Listing}
\usepackage{color} 

\input{math_commands.tex}

\usepackage{subfig}
\usepackage{multirow}
\usepackage{booktabs}

\title{Adaptive Policy Learning for Offline-to-Online Reinforcement Learning}
\author{
    Han Zheng\textsuperscript{\rm 1}\thanks{This work was done when Han Zheng was interning with Microsoft Research Asia.},
    Xufang Luo\textsuperscript{\rm 2}\thanks{Corresponding author. Email: xufluo@microsoft.com},
    Pengfei Wei\textsuperscript{\rm 3},\\
    Xuan Song\textsuperscript{\rm 4},
    Dongsheng Li\textsuperscript{\rm 2},
    Jing Jiang\textsuperscript{\rm 1}
}
\affiliations{
    \textsuperscript{\rm 1}University of Technology Sydney,
    \textsuperscript{\rm 2}Microsoft Research Asia,
    \textsuperscript{\rm 3}National University of Singapore,\\
    \textsuperscript{\rm 4}Southern University of Science and Technology

}
\begin{document}

\maketitle

\begin{abstract}
Conventional reinforcement learning (RL) needs an environment to collect fresh data, which is impractical when online interactions are costly.
Offline RL provides an alternative solution by directly learning from the previously collected dataset. However, it will yield unsatisfactory performance if the quality of the offline datasets is poor.
In this paper, we consider an offline-to-online setting where the agent is first learned from the offline dataset and then trained online, and propose a framework called Adaptive Policy Learning for effectively taking advantage of offline and online data.
Specifically, we explicitly consider the difference between the online and offline data and apply an adaptive update scheme accordingly, that is, a pessimistic update strategy for the offline dataset and an optimistic/greedy update scheme for the online dataset.
Such a simple and effective method provides a way to mix the offline and online RL and achieve the best of both worlds. 
We further provide two detailed algorithms for implementing the framework through embedding value or policy-based RL algorithms into it.
Finally, we conduct extensive experiments on popular continuous control tasks, and results show that our algorithm can learn the expert policy with high sample efficiency even when the quality of offline dataset is poor, e.g., random dataset.
\end{abstract}

\section{Introduction}\label{section:intro}

Conventional online reinforcement learning (RL) methods~\citep{sac,td3} usually learn from experiences generated by online interactions with the environment.
They are impractical in some real-world applications, e.g., dialog~\citep{dialog} and education~\citep{education}, where interactions are costly.
Furthermore, many offline data have already been generated by one or more policies, but online RL agents usually cannot utilize them directly~\citep{Scottbcq}.
In contrast, supervised learning achieves remarkable successes across a range of domains~\citep{sl,gpt3} by directly leveraging existing large-scale datasets, such as ImageNet~\citep{ImageNet}.
To apply such data-driven learning paradigm with RL objectives, many offline RL methods have been proposed and aroused much attention recently~\citep{Levine2020OfflineRL}.  
They try to learn policies from static datasets which are pre-collected by arbitrary policies.

Existing offline RL studies mainly focus on remedying the issue of distribution mismatch or out-of-distribution (OOD) actions by employing a pessimistic update scheme~\citep{bear,cql} or in combination with imitation learning~\citep{Scottbcq}.
However, when the dataset is fixed and sub-optimal, it is nearly impossible for offline RL to learn the optimal policy~\citep{Kidambi2020MOReLM}.
Even worse, when a large portion of actions or states is not covered within the training set distribution, offline RL methods usually fail to learn good policies~\citep{cql,fu2020d4rl,Levine2020OfflineRL}. 
Moreover, policy evaluation under the offline RL setting is also challenging. Although some methods, such as off-policy evaluation algorithms~\citep{opesurvey}, are proposed for this problem, they are still not ideal for the practical purpose.  

Some recent works address the above issues by employing an offline-to-online setting.
Such methods~\citep{lee2021offlineonline,awac} focus on pre-training a policy using the offline dataset and fine-tuning the policy through further online interactions.
Even though their methods alleviate the above problems, they have not considered the different advantages of offline and online data and utilised both well.
Specifically, offline data can prevent agents from prematurely converging to sub-optimal policies thanks to the potential data diversity, while online data can stabilize training and accelerate convergence~\citep{awac,sutton2011reinforcement}.
They can contribute to better policy learning in different ways.
But prior methods only stress one of them.
For instance,~\citet{awac} employs a pessimistic strategy to update the policy, which may be problematic for improving policy performance during the online phase.
Unlike~\citet{awac},~\citet{lee2021offlineonline} only selects near-on-policy data from the offline dataset during online phase to alleviate the distribution shift in transition from offline to online.
Such a strategy ignores a large part of the offline dataset, whose potential data diversity is important for learning better policies.

To tackle the above problems, in this paper, we propose that \textit{the advantage of online and offline data should be emphasized in an adaptive way.}
First, considering their different characteristics, separate updating strategies should be employed for online and offline data, respectively.
To do so, we present a novel framework called \textit{Adaptive Policy Learning} (APL) that takes advantages of them effectively.
The core idea is simple: when learning from online data, the agent is updated in an optimistic way, and when learning from the offline dataset, the agent is optimized by a pessimistic strategy. The intuition behind this is that near-on-policy data can be collected via online interactions, so an optimistic strategy is used here for better policy improvement, and potentially useful offline datasets can be collected by arbitrary policies, so a pessimistic strategy is used to exploit all these data.
In addition, to distinguish between offline and online data in a simple way, we design a two-level replay buffer for the APL framework.
We further provide value-based and policy-based implementations for APL framework through embedding state-of-the-art (SOTA) online and offline RL methods into the framework.
Experimental results demonstrate that our algorithm can learn expert policies in most tasks with a high sample efficiency regardless of the quality of offline datasets.
More specifically, our algorithm performs better by using only $20\%$ online interactions compared with the previous offline-to-online method~\citep{awac}.

Our contributions can be summarized as below:
\begin{itemize}
\item We propose a simple framework called Adaptive Policy Learning (APL) that considers different advantages of offline and online data for policy learning, and can effectively make use of them. 
\item We further provide value-based and policy-based algorithms to implement the APL framework, showing APL is compatible with various RL methods.
\item We test our algorithm on the popular continuous control tasks MuJoCo~\citep{todorov2012mujoco} and compare it with some strong baselines. The results clearly show that the APL framework is effective under the offline-to-online setting.
\end{itemize}

\section{Related Work}
\textbf{Online RL} In general, online RL algorithms can be divided into two categories, i.e., on-policy and off-policy algorithms. On-policy methods~\citep{trpo,ppo} update the policy using data collected by its current behavior policy. 
As ignoring the logged data collected by its history behavior policies, they usually have a lower sample efficiency than the off-policy methods~\citep{td3,redq}, which enable the policy to learn from experience collected by history behavior policies. 
However, they cannot learn from history trajectories collected by other agents' behavior policies~\citep{Scottbcq,cql}.
So, the need for huge online interaction makes online RL impractical for some real-world applications, such as dialog agents~\citep{dialog} or the education system~\citep{education}.

\textbf{Offline RL}
Offline RL algorithms assume the online environment is unavailable and learn policies only from the pre-collected dataset. 
As the value estimation error cannot be corrected using online interactions, these methods utilize a pessimistic updating strategy to relieve the distribution mismatch problem~\citep{Scottbcq, bear}. 
Model-free offline RL methods generally employ value or policy penalties to constrain the updated policy close to the data collecting policy~\citep{Wu2019BehaviorRO,cql,Scottbcq,popo,ghosh2022offline}.
Model-based methods use predictive models to estimate uncertainties of states and then update the policy in a pessimistic way based on them~\citep{Kidambi2020MOReLM,Yu2020MOPOMO,chen2021offline}.
Those offline RL methods cannot guarantee a good performance, especially when the data quality is poor~\citep{cql}. 
Besides, policy evaluation when the online environment is unavailable is also challenging. Even though off-policy evaluation (OPE) methods~\citep{opesurvey} present alternative solutions, they are still far from perfect. 

The above issues in online and offline RL motivate us to investigate the offline-to-online setting.

\textbf{Offline-to-online RL}
Some works focus on the mixed setting where the agent is first learned from the offline dataset and then trained online. 
\citet{awac} propose an advantage-weighted actor-critic (AWAC) method that restricts the policy to select actions close to those in the offline data by an implicit constraint.
\citet{iql} present IQL, implementing a weighted behavioral cloning step for better online policy improvement, which can also be used in offline-to-online setting.
When online interactions are available, such conservative designs may adversely affect the performance.
OFF2ON~\cite{lee2021offlineonline} employs a balanced replay scheme to address the distribution shift issue. It uses offline data by only selecting near-on-policy samples.
Unlike these works, our method addresses all online and offline data, and explicitly considers the difference between them by adaptively applying optimistic and pessimistic updating schemes for online and offline data, respectively.
Besides, our framework is more flexible because it can be easily applied to policy or value-based methods.
Moreover, our algorithm is much more robust to different hyper-parameters (see the experiment section).
Particularly, without careful tuning, our algorithm can achieve better or comparable performance than prior methods, while OFF2ON uses a specialized network architecture, and the hyper-parameters are fine-tuned.
\citet{Deployment-efficient} focus on optimizing deployment efficiency, i.e., the number of distinct data-collection policies used during learning, by employing a behavior-regularized policy updating strategy.
However, they ignore existing offline dataset, and dose not focus on improving sample efficiency, while both are addressed in our paper.
Some works~\citep{zhu2019dexterous,vecerik2017leveraging,rajeswaran2017learning,Kim2013LearningFL} can also learn from online interactions and offline data. However, they need expert demonstrations instead of any dataset, limiting their applicability.

\section{Preliminaries}
In RL, interactions between the agent and environment are usually modelled using Markov decision process (MDP) $\left(\mathcal{S}, \mathcal{A}, p_{\text{M}}, r, \gamma\right)$, with state space $\mathcal{S}$ (state $\mathbf{s}\in\mathcal{S}$), action space $\mathcal{A}$ (action $\mathbf{a}\in\mathcal{A}$).
At each discrete time step, the agent takes an action $\mathbf{a}$ based on the current state $\mathbf{s}$, and the state changes into $\mathbf{s}'$ according to the transition dynamics $p_{\text{M}}\left(\mathbf{a}^{\prime} \mid \mathbf{s}, \mathbf{a}\right)$, and the agent receives a reward $r \in \mathbb{R}$. The agent's objective is to maximize the return, which is defined as $R_{t}=\sum_{n=0}^{\infty} \gamma^{n} r_{t+n}$, where $t$ is the time step, and $\gamma \in[0,1)$ is the discounted factor.
The mapping from $\mathbf{s}$ to $\mathbf{a}$ is denoted by the stochastic policy $\pi: \mathbf{a}\sim \pi(\cdot \vert \mathbf{s})$. Policy can be stochastic or deterministic, and we use the stochastic from in this paper for generality.
Each policy $\pi$ have a corresponding action-value function $Q^{\pi}(\mathbf{s}, \mathbf{a})=\mathbb{E}_{\pi}\left[R_{t} \mid \mathbf{s}, \mathbf{a}\right]$, which is the expected return following the policy after taking action $\mathbf{a}$ in state $\mathbf{s}$.
The action-value function of policy $\pi$ can be updated by the Bellman operator $\mathcal{T}^{\pi}$:
\begin{equation}
\mathcal{T}^{\pi} Q(\mathbf{s}, \mathbf{a})=\mathbb{E}_{\mathbf{s}^{\prime}}\left[r+\gamma Q\left(\mathbf{s}^{\prime}, \pi\left(\mathbf{s}^{\prime}\right)\right)\right]
\end{equation}
Q-learning~\citep{sutton2011reinforcement} directly learns the optimal action-value function $Q^{*}(\mathbf{s}, \mathbf{a})=\max _{\pi} Q^{\pi}(\mathbf{s}, \mathbf{a})$, and such Q-function can be modelled using neural networks~\citep{dqn}.

In principle, off-policy methods, such as Q-learning, can utilize experiences collected by any policy, and thus they usually maintain a replay buffer $\mathcal{B}$ to store and repeatedly learn from experiences collected by behavior policies~\citep{agarwal2020optimistic}.
Such capability also enables off-policy methods in the offline setting by storing offline data into the buffer $\mathcal{B}$ and not updating the buffer during learning since no further interactions are available here~\citep{Levine2020OfflineRL}.
However, this simple adjusting generally results in poor performance due to bootstrapping errors from OOD actions, especially when the dataset is not diverse~\citep{cql,agarwal2020optimistic,Scottbcq}, and this is also the problem tackled in most offline RL works.

This paper focuses on the offline-to-online setting, where the agent learns from both the offline and online datasets. 
We use off-policy methods in the online part because they can utilize data more effectively and generally have higher sample efficiency than on-policy ones. Thus, if no additional marks, online RL methods refer to off-policy algorithms in the rest of this paper.

\begin{figure}[t]
\centering
\includegraphics[width=0.4\textwidth]{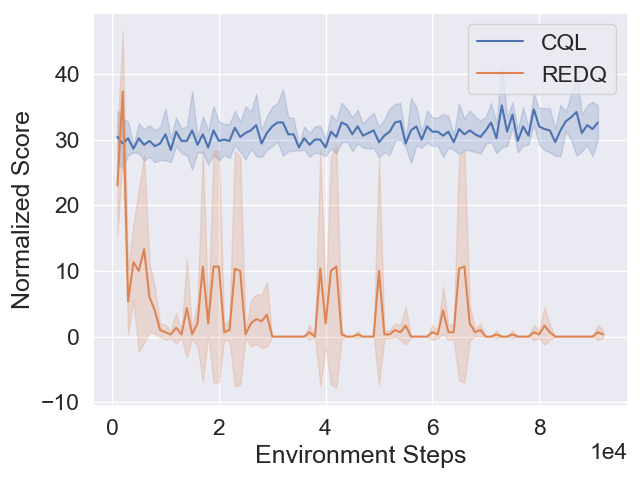}
\caption{Learning curves for the online agent initialized with the D4RL~\citep{fu2020d4rl} dataset hopper-medium-replay-v0. Scores are averaged over five random seeds and the shaded areas represent the standard deviation.
The normalized score of 100 is the average returns of a domain-specific expert while normalized score of 0 corresponds to the average returns of an agent taking uniformly random actions across the action space.}
\label{fig:hopper_example}
\end{figure}

\section{Methodology}

In this section, we first give an illustrative example to explain our motivation.
Then, we introduce the proposed APL framework, trying to couple online and offline RL in an adaptive way. 
Finally, we present detailed algorithms for implementing APL framework.

\textbf{An illustrative example} We test the SOTA offline and online RL methods under the offline-to-online setting, i.e., Conservative Q-learning (CQL)~\citep{cql} and Randomized Ensembled Double Q-learning (REDQ)~\citep{redq}, respectively.
Specifically, we first pre-train the agent with the offline dataset for $100K$ steps. Then, the agent is fine-tuned online by alternately conducting the interaction and updating process. The agent interacts with the environment for $1K$ steps, and is updated for $10K$ steps.
The total online interaction steps are around $100K$.

The offline-to-online setting described above should be a favourable one for policy learning because both diverse offline data and online interactions are available. Therefore, we expect the following results: the initial offline training with the fixed dataset will provide relatively good but not expert performance for the agent, 
and then the agent can be further improved by the online process since online interaction data can be obtained.
At last, we may acquire a well-performed policy, with a normalized score close to or better than $100$ (i.e., the performance of an expert).

Nonetheless, the experiment results in Figure \ref{fig:hopper_example} do not show what we expected.
Specifically, starting points of the two curves show that the offline algorithm CQL and online algorithm REDQ can obtain normalized scores greater than $0$ (i.e., the performance of a random policy).
This indicates that by learning from the offline dataset, the agent can obtain a relatively good starting point, as expected. And these starting scores (below $30$) are also far from the expert score, which leaves a large room for further promotion in the online phase.
However, both algorithms have trouble in the online process.
REDQ suffers from severe instability, and in the end, the policy almost degenerates into a random one. 
Although the CQL agent can keep stable during learning, its improvement is very limited, which indicates it is ineffective for leveraging online interaction.  

These results indicate that a pure online RL algorithm may be problematic for handling offline and online data in a single training process. Furthermore, the pure offline RL algorithm cannot effectively use valuable online interaction data due to its conservative updating strategy. 

\subsection{Adaptive Policy Learning Framework}

In this subsection, we present the \textit{Adaptive Policy Learning} (APL) framework, trying to tackle the above problem by adaptively leveraging online and offline data.
The underlying idea is simple and can be described as follows.
When online interactions are available, we try to obtain near-on-policy data from them and choose an optimistic updating strategy since these data reflect the true situation of the current policy.
By contrast, we use a more pessimistic updating strategy when data are sampled from the offline dataset.
In this way, we can make full use of both online and offline data and explicitly consider their differences by separately applying suitable updating schemes.
It is worth stressing that being optimistic here means we rely on an online RL update strategy which is optimistic compared with offline RL methods, while being pessimistic means an offline RL update strategy.

Then, we give a general formalization for the above idea. Since this idea can be applied to both value-based and policy-based methods, we use $C$ to represent a policy or state-action value function, and define a unified updating rule as follows:
\begin{equation}
\begin{aligned}
\label{eqn:aclf}
C^{k+1} \leftarrow \mathcal{F}\left(\mathbb{A}(C^k) + \mathcal{W}(\mathbf{s},\mathbf{a}) \mathbb{B}(C^k)\right),
\end{aligned}
\end{equation}
where $k$ is the times of iterative updates.
For the right-hand side, the first term $\mathbb{A}(C)$ stands for the optimistic updating strategy. It is the common learning target used in online RL, such as the Bellman error for value-based methods.
The second term $\mathbb{B}(C)$ stands for the pessimistic updating strategy. It is a penalty term to make the learned agent take actions close to the dataset or take conservative actions.
Besides, to express the idea of adaptive learning, we apply a weight function $\mathcal{W}(\mathbf{s},\mathbf{a})$ to the penalty term $\mathbb{B}(C)$.
Specifically, when we use near-on-policy data, $\mathcal{W}(\mathbf{s},\mathbf{a})$ will be a small value, and the updating relies more on $\mathbb{A}(C)$, leading to a relatively optimistic updating strategy.
On the contrary, when we use the offline data, $\mathcal{W}(\mathbf{s},\mathbf{a})$ will be a large value, and the updating strategy is relatively pessimistic.
In addition, we use $\mathcal{F}$ to denote the general updating operator, such as $\operatorname{argmin}$ or $\operatorname{argmax}$. 
Using this general operator, the updating rule for both value and policy-based methods can be unified in Eq.~\ref{eqn:aclf}. Note that we only introduce the general framework here, and detailed algorithms are presented in the next sections.

\textbf{Online-Offline Replay Buffer}
\label{sec:oorb}
We now introduce a simple but effective online-offline replay buffer (OORB) to distinguish between near-on-policy and offline data.
OORB consists of two buffers.
One is the online buffer that collects the near-on-policy data generated via online interactions. 
To ensure the data in the online buffer is near-on-policy, we set this buffer to be small, and newly collected online data are stored in it by following the first-in-first-out rule.
The other is the offline buffer containing previously collected offline datasets which can come from arbitrary policies, and all data generated via interactions in the online phase.

Data are sampled from OORB following a Bernoulli distribution.
With a probability $p$, data are sampled from the online buffer, and with probability $1-p$, they are sampled from the offline buffer.
The effect of $p$ on the final performance is tested via ablation studies in the experiment section.

\begin{algorithm}[tb]
 \caption{Greedy Conservative Q-ensemble Learning}
  \label{alg:our}
\begin{algorithmic}
  \STATE // Online interaction steps in each iteration $T_\text{on}$, updating steps in each iteration $T_\text{off}$, initial offline training steps $T_\text{initial}$
  \STATE // OORB threshold $p$, OORB starting sampling size $T_\text{s}$
  \STATE {\bfseries Input:} Total online interaction steps $S_T$
  \STATE Initialize online buffer $B_{\text{on}}$ to empty, offline buffer $B_{\text{off}}\gets \text{offline dataset}$ 
  \STATE Initialize policy $\pi$, Q-functions $Q_{i}, i\in N$
  \STATE Set the updating step counter $t$ 
  \STATE Set the total online interaction step counter $S_{\text{on}}\gets 0$
  \STATE Train the agent for $T_\text{initial}$ steps using the offline dataset
  \REPEAT
  \STATE $t \gets 0 $
  \STATE Interact with the environment for $T_\text{on}$ steps
  \STATE Store collected experiences to both online buffer $B_{\text{on}}$ and offline buffer $B_{\text{off}}$
  \STATE $S_\text{on}\gets S_\text{on}+T_\text{on}$
  \FOR{ $t < T_\text{off}$}
  \STATE Sample a random value $p_s\sim\mathbb{U}(0,1)$
  \IF{$p_s<p$ and $S_\text{on}>T_\text{s}$}
  \STATE Sample a batch of $(\mathbf{s},\mathbf{a})$ from online buffer $B_{\text{on}}$
  \STATE Set the $\mathcal{W}(\mathbf{s},\mathbf{a})$ to 0
  \ELSE
  \STATE Sample a batch of $(\mathbf{s},\mathbf{a})$ from offline buffer $B_{\text{off}}$
  \STATE Set the $\mathcal{W}(\mathbf{s},\mathbf{a})$ to 1
  \ENDIF
  \STATE Update the Q-functions $Q_{i}, i\in N$ by Eq. \ref{eqn:gcql_q}
  \STATE Update the policy $\pi$ by Eq. \ref{eqn-redq:policy-update}
  \STATE $t\gets t+1$
  \ENDFOR
  \UNTIL{$S_\text{on} > S_T$}
  \end{algorithmic}
\end{algorithm}

\subsection{Value-Based Implementation}
\label{sec:algorithms}

We first present an action-value based implementation for APL framework by incorporating CQL~\citep{cql} as the value-based offline RL method and an ensemble online RL algorithm REDQ~\citep{redq}.
We name the implementation Greedy-Conservative Q-ensemble Learning (GCQL).

Specifically, $C$ in Eq.~\ref{eqn:aclf} is the state-action value $Q$ here, and we use the updating function in REDQ as the optimistic strategy. Since REDQ is an ensemble method which uses a set of Q-functions, we use $i$ as the index of the Q-functions, and the size of the set is $N$. Hence,
\begin{equation}
\small
\begin{aligned}
\label{eqn:acqe}
 \mathbb{A}(Q^k_i)= \mathbb{E}_{\mathbf{s},\mathbf{a},\mathbf{s'}\sim{\text{OORB}}, \mathbf{a'}\sim\pi^k(\cdot \mid s')}\left[\left(Q^k_i(\mathbf{s}, \mathbf{a})-\mathcal{B}^{\pi} \hat{Q}^{k}(\mathbf{s'}, \mathbf{a'})\right)^{2}\right],
\end{aligned}
\end{equation}
where $k$ is the times of iterative updates same as Eq.~\ref{eqn:aclf}, and ``$\sim{\text{OORB}}$" represents sampling data from OORB. 
Operator $\mathcal{B}^{\pi}\hat{Q}^{k}(\mathbf{s'}, \mathbf{a'})$ is
\begin{equation}
r+\gamma \min _{i \in \mathcal{M}} \hat{Q}^k_{i}\left(\mathbf{s'}, \mathbf{a'}\right), \quad \mathbf{a'} \sim \pi^k\left(\cdot \mid \mathbf{s'}\right).
\end{equation}
Here, $\pi^k$ is the current policy, and $\hat{Q}$ denotes a target $Q$ function for stabilizing the learning process~\citep{dqn}. Following REDQ's design, we randomly select two Q-functions from a set of them for ensemble, with $\mathcal{M}$ representing the set of selected indexes.

Then, we adopt the conservative regularizer in CQL as the second term in Eq.~\ref{eqn:aclf}. For every Q-function in the set:
\begin{equation}
\small
\begin{aligned}
\label{eqn:gcql_b}
 \mathbb{B}(Q^k_i)= \alpha \mathbb{E}_{\mathbf{s}\sim{\text{OORB}}}\left[\log \sum_{\mathbf{a'}} \exp (Q^k_i(\mathbf{s}, \mathbf{a'}))-\mathbb{E}_{\mathbf{a} \sim {\text{OORB}}}[Q^k_i(\mathbf{s}, \mathbf{a})]\right],
\end{aligned}
\end{equation}
where action $\mathbf{a'}$ is sampled from the current policy, i.e., $\mathbf{a'} \sim \pi^k(\cdot|s)$. 
Hence, the overall updating function of $Q$ is:
\begin{equation}
\label{eqn:gcql_q}
Q^{k+1}_i=\arg \min _{Q^k_i} \left\{\mathbb{A}(Q^k_i)+\mathcal{W}(\mathbf{s},\mathbf{a})\mathbb{B}(Q^k_i)\right\}
\end{equation}

Finally, the update function of policy is shown as follows:
\begin{equation}
\small
\label{eqn-redq:policy-update}
\begin{aligned}
\pi^{k+1}=\arg \max _{\pi^k}\mathbb{E}_{\mathbf{a} \sim \pi^k\left(\cdot \mid \mathbf{s}\right)}\left[ \mathbb{E}_{i\in N}\left[Q_i^k\left(\mathbf{s}, \mathbf{a}\right)\right]-\alpha \log \pi^k\left(\mathbf{a} \mid\mathbf{s}\right)\right].
\end{aligned}
\end{equation}

\subsection{Policy-Based Implementation}\label{sec:gctd3bc} 

Besides value-based methods, APL framework can also be easily implemented with policy-based methods. Here we take TD3BC~\citep{td3bc} as an example, and name our algorithm Greedy-Conservative TD3BC (GCTD3BC). Specifically, TD3BC uses the policy update step in TD3, and adds a behavior cloning (BC) term to regularize the policy. Therefore, GCTD3BC takes the policy update function in TD3 as the optimistic term, and takes the BC regularizer as the pessimistic term. So, the policy update rule in GCTD3BC is
\begin{equation}
\label{eqn:gctd3bc}
\begin{aligned}
\pi^{k+1}=\underset{\pi^k}{\operatorname{argmax}} \mathbb{E}_{(\mathbf{s}, \mathbf{a}) \sim \text{OORB}}\left[\mathbb{A}(\pi^k)+\mathcal{W}(\mathbf{s},\mathbf{a})\mathbb{B}(\pi^k)\right],
\end{aligned}
\end{equation}
where $\mathbb{A}(\pi^k)=\lambda Q^k(\mathbf{s},\pi(\mathbf{s}))$ is the value function in TD3, and $\mathbb{B}(\pi^k)=-(\pi^k(\mathbf{s})-\mathbf{a})^{2}$ is the BC regularizer. Hyper-parameter $\lambda$ is used to adjust two terms.

We now introduce how $\mathcal{W}(\mathbf{s},\mathbf{a})$ is set in both implementations.
We use a very simple method: when data is sampled from the online buffer, $\mathcal{W}(\mathbf{s},\mathbf{a})$ is set to 0, otherwise 1.
This is why we use the term \textit{Greedy-Conservative} to describe our algorithms.
Specifically, we greedily exploit the near-on-policy data with the online RL scheme without any conservative regularizer. And we conservatively exploit the offline data by employing the offline RL methods.
Formally, this strategy can be explained as below:
\begin{equation}
\label{eq:w}
\mathcal{W}(\mathbf{s},\mathbf{a}) \gets
  \begin{cases}
    0  &~\text{if}~(\mathbf{s},\mathbf{a})\sim\text{online buffer}\\
    1  &~\text{otherwise}
  \end{cases}
\end{equation}

To be more clear, we summarize GCQL in Algorithm~\ref{alg:our}, and explain the main steps as follows.
Firstly, we learn from the existing offline data for $T_{\text{initial}}$ steps to leverage them. 
To make good use of the offline data, we usually set $T_\text{initial}$ to a large value, e.g., $100K$. Secondly, we begin the following interleaving learning process. We conduct the online interaction for $T_\text{on}$ steps and store newly collected experiences to OORB, and then we update the agent for $T_\text{off}$ steps. For higher sample efficiency, we set the number of online interaction steps $T_\text{on}$ to a small value, e.g., $1K$, and the number of updating steps $T_\text{off}$ to a value larger than $T_\text{on}$, e.g., $10K$.
If the batch of data for updating the policy and Q-functions comes from the online buffer, $\mathcal{W}(\mathbf{s},\mathbf{a})$ is set to 0, otherwise 1.

Algorithm GCTD3BC is similar to Algorithm~\ref{alg:our} except that GCTD3BC uses the action-value objective in TD3~\citep{td3} and the policy updating rule is defined in Eq. \ref{eqn:gctd3bc}.
\begin{table*}[t]
  \vskip 0.15in
  \centering
  \resizebox{\textwidth}{!}{%
  \begin{tabular}{lllllllllll}
    \toprule
    Environment & GCQL & GCTD3BC & CQL & REDQ\_ON& REDQ & TD3\_ON & TD3BC &AWAC& OFF2ON & IQL\\
    \midrule
    walker2d-r & 31$\pm$27 & 5$\pm$3 & 7$\pm$9 & \textbf{71$\pm$11} & 5$\pm$3 &7$\pm$2& 6$\pm$3 & 12 & 20$\pm$13 & 7$\pm$3\\
    hopper-r & 58$\pm$30 & 35$\pm$22 & 10$\pm$1 & 78$\pm$37 & 2$\pm$1 &10$\pm$2& 11$\pm$0& 63 & \textbf{81$\pm$21} & 10$\pm$2 \\
    halfcheetah-r& \textbf{101$\pm$2} & 69$\pm$8 & 46$\pm$4 &59$\pm$2& 32$\pm$1&39$\pm$1 & 35$\pm$3 & 53 & 85 $\pm$3 &28$\pm$7 \\
    \midrule
    walker2d-m & \textbf{94$\pm$6} & 90$\pm$7 & 83$\pm$1 & 71$\pm$11 &2$\pm$3&7$\pm$2 &79$\pm$2& 80 & 89$\pm$2 &51$\pm$13 \\
    hopper-m & 83$\pm$11 & \textbf{99$\pm$3} & 70$\pm$23 &78$\pm$37&3$\pm$1 &10$\pm$2&80$\pm$13&91 & 59 $\pm$9 &42$\pm$9 \\
    halfcheetah-m & \textbf{66$\pm$3} & 62$\pm$2 & 25$\pm$8 &59$\pm$2&46$\pm$1 &39$\pm$1&43$\pm$1 & 41 & 58$\pm$2&40$\pm$0 \\
    \midrule
    walker2d-me & 93$\pm$12& 102$\pm$2 & 105$\pm$1 &71$\pm$11& 12$\pm$3 &7$\pm$2&\textbf{110$\pm$3}& 78 & 101$\pm$24 &58$\pm$22 \\
    hopper-me & \textbf{110$\pm$1} & \textbf{110$\pm$2} & 109$\pm$5 &78$\pm$37&40$\pm$15 &10$\pm$2&\textbf{110$\pm$0}& \textbf{112} & 82 $\pm$21 & 72$\pm$16 \\
    halfcheetah-me & \textbf{102$\pm$1} & \textbf{103$\pm$2} & 92$\pm$2 &59$\pm$2&9$\pm$3 &39$\pm$1&98$\pm$2& 41 &100$\pm$1 &38$\pm$17 \\
    \midrule
    walker2d-mr & \textbf{97$\pm$16}& 90$\pm$9 & 57$\pm$5 &71$\pm$11&13$\pm$2 &7$\pm$2&60$\pm$5& - &71$\pm$32 & 30$\pm$13\\
    hopper-mr & 72$\pm$20& \textbf{87$\pm$11} & 37$\pm$4 &78$\pm$37&0$\pm$1 &10$\pm$2&38$\pm$1& - &  \textbf{60$\pm$23} & 31$\pm$10\\
    halfcheetah-mr & \textbf{62$\pm$2}& 53$\pm$1 &50$\pm$0 &\textbf{59$\pm$2}&28$\pm$25 &39$\pm$1&47$\pm$0& - &57$\pm$1 &42$\pm$2\\
    \midrule
    Total &\textbf{969$\pm$131} & 905$\pm$71 & 691$\pm$63 &624$\pm$200&192$\pm$59 &224$\pm$20&717$\pm$33 & - & 863$\pm$152 &449$\pm$114\\
    \bottomrule
  \end{tabular}
  }%
  \caption{Averaged normalized score over last three evaluation iterations and 5 random seeds. 
  $\pm$ captures the standard deviation over seeds. The highest performing scores are in bold. 
  The results of AWAC are taken from their paper \cite{awac}.
  REDQ\_ON and TD3\_ON are pure online methods without any offline pre-training. We rerun them using implementations from authors to ensure identical evaluation process.
  Suffix ``-r" = random-v0 dataset, ``-m" = medium-v0 dataset, ``-me" = medium-expert-v0, and ``-mr" = medium-replay-v0. All learning curves are showed in Appendix.}
  \label{table: last_performance}
\end{table*}

\section{Experiments}
In this section, we design experiments to verify the effectiveness of our framework from three perspectives:
(1) the performance compared with competitive baselines;
(2) ablation studies to test the effect of key components used in our methods;
(3) the influence of different hyper-parameters.

\textbf{Tasks} All experiments were done on the continuous control task set MuJoCo~\citep{todorov2012mujoco}, and the offline dataset comes from the popular offline RL benchmark D4RL~\citep{fu2020d4rl}. Here, we test three tasks, i.e., walker2d, hopper and halfcheetah, and each task takes four different kinds of offline dataset, which are random-v0, medium-v0, medium-replay-v0 and medium-expert-v0.

\textbf{Settings} We set $T_{\text{on}}$ in Algorithm~\ref{alg:our} to $1K$. To better exploit the offline dataset, we set $T_{\text{initial}}$ and $T_\text{off}$ to $100K$ and $10K$, respectively.
For OORB, we set $p=0.5$ for GCQL and $p=0.1$ for GCTD3BC, and $T_\text{s}$ to $10K$ for both of them. The size of online and offline buffer is set to $20K$ and $3M$, respectively.
We input $S_T$ as $100K$.
The above configurations keep the same across all tasks, datasets and methods.

\begin{figure*}[!ht]
\centering
\includegraphics[width =\textwidth]{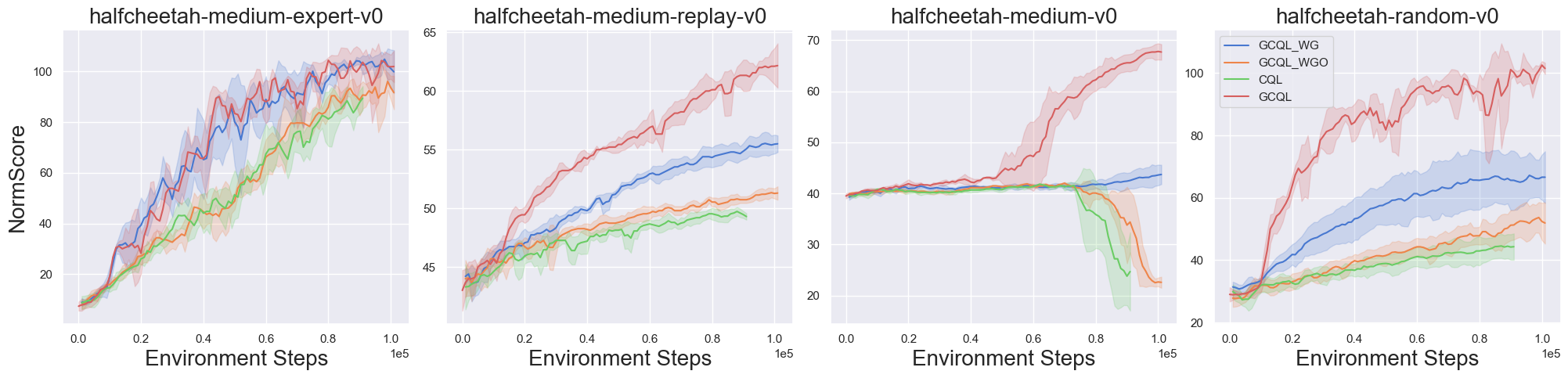}
\caption{Ablation studies on Halfcheetah. GCQL\_WG: GCQL without the greedy updating strategy. GCQL\_WGO: GCQL without the greedy updating strategy and online buffer. Results on four tasks are averaged over three random seeds.} 
\label{fig:halfcheetah_ab}
\end{figure*}

\textbf{Compared Methods }
We compare our algorithms, i.e., GCQL and GCTD3BC, with 8 competitive baselines, i.e., CQL, REDQ\_ON, REDQ, TD3\_ON, TD3BC, AWAC~\cite{awac}, OFF2ON~\cite{lee2021offlineonline} and IQL~\cite{iql}. REDQ\_ON denotes that we rerun the officially released codes by the authors~\citep{redq} without changing hyper-parameters, and do not use any offline data to pre-train for this online method. We use the ``\_ON" to indicates the agent is trained purely online without offline pre-training, and methods without ``\_ON" are in the offline-to-online setting. Similarly, TD3\_ON is a pure online agent trained with TD3. For a fair comparison, we also include a re-implemented version in GCQL, i.e. REDQ, as a baseline, where the number of Q-functions is set to 5 for computational efficiency (as in GCQL) and pre-training with offline data is leveraged. The results of AWAC are directly taken from their paper. For OFF2ON, since the authors did not release the code for pre-training, we contact the authors and use pre-trained agents provided by them, and then use their codes for the rest of training. For other methods, results are obtained by rerunning their codes under the offline-to-online setting. We use default hyper-parameters for all methods without further tuning.

\subsection{Overall Performance}\label{sec:operformance}
We list scores in Table~\ref{table: last_performance}, and include all learning curves in Appendix. First, it is clear that our methods GCQL and GCTD3BC perform much better than baselines, well demonstrating the effectiveness of them. Second, we notice that our methods are more robust comparing with baselines. Specifically, offline RL methods (i.e., CQL and TD3BC) can obtain higher scores than pure online methods (i.e., REDQ\_ON and TD3\_ON) when the dataset is not random, but they have poor performance when the dataset is random. In contrast, offline-to-online methods (e.g., GCQL and OFF2ON) can perform well regardless of random dataset or not. Note that OFF2ON is fine-tuned. For instance, the critic's network architecture is different from that in the original CQL paper. Our methods use default hyper-parameters in the original paper without fine-tuning, and can perform better than those fine-tuned methods. Third, it is surprising that single-agent method GCTD3BC have higher scores and faster learning speed than ensemble method GCQL on some tasks, such as hopper-medium-v0. All results shows APL can benefit more from the online and offline data and gain high sample efficiency. Also, the general APL framework can be successfully applied to the various RL methods.

\begin{figure*}[!ht]
\centering
\includegraphics[width =\textwidth]{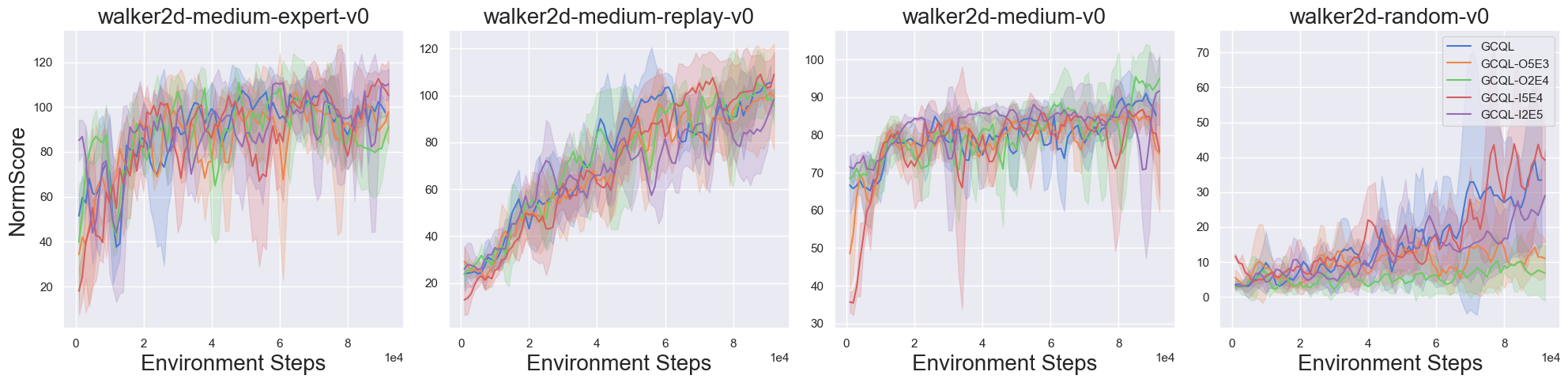}
\caption{
GCQL-i2e5 and GCQL-i5e4 mean that $T_\text{initial}$ is $2e5$ and $5e4$, respectively. And GCQL-o2e4 and GCQL-o5e3 mean $T_\text{off}$ is set to corresponding values. Results are averaged over three random seeds.}
\label{fig:io}
\end{figure*}

\begin{figure*}[!ht]
\centering
\includegraphics[width =\textwidth]{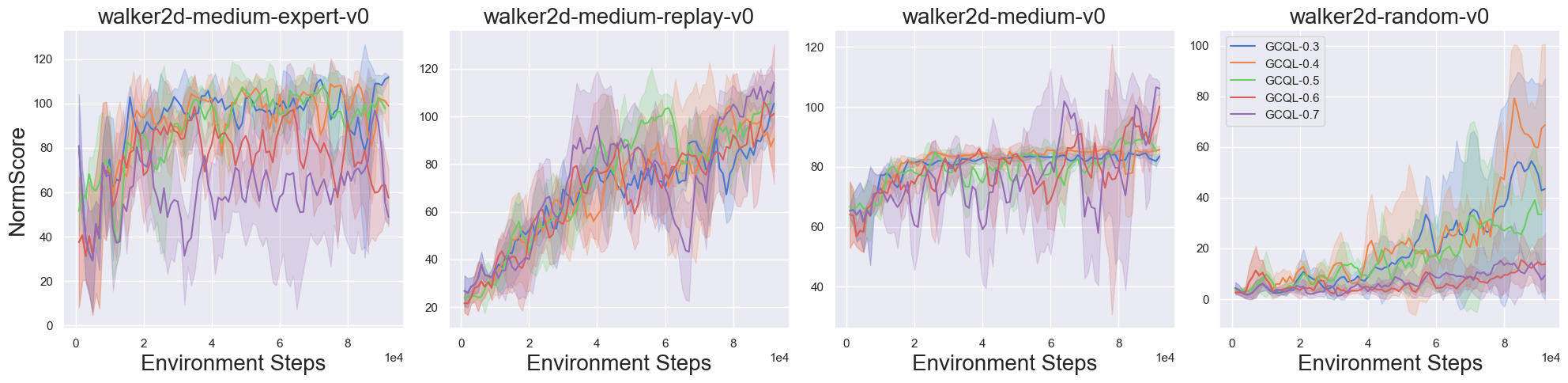}
\caption{Learning curves for agents with different OORB threshold $p$. GCQL-X means that $p$ is set to X. Results are averaged over three random seeds.}
\label{fig:p}
\end{figure*}

\subsection{Ablation Study on Key Components in APL}\label{sec:ab}
To investigate the effect of key components in our method, we conduct the following ablation studies on GCQL.
The main differences between APL framework and prior offline RL methods is the adaptive update schema and online buffer in OORB.
To this end, we design two variants of GCQL to investigate such two components' effect.
\textbf{GCQL\_WG}: We remove the adaptive update schema by fixing $\mathcal{W}$ to 1, resulting in the variant of GCQL \textbf{w}ithout \textbf{g}reedy updating strategy.
\textbf{GCQL\_WGO}: We further remove the online buffer, resulting in the variant of GCQL \textbf{w}ithout \textbf{g}reedy updating strategy and \textbf{o}nline buffer.

It is easy to deduce that when the dataset includes the expert data, all variants perform well.
The underlying reason is obvious as offline RL methods can perform quite well when data quality is good.
However, for other datasets with worse quality (e.g., medium-replay, medium and random), online buffer and greedy update schema play an important role in boosting the performance and stabilizing the learning process.
Results show that for these worse datasets, both GCQL\_WG and GCQL\_WGO have a clear performance drop compared with GCQL.
Specifically, it is observed that, for medium datasets, both CQL and GCQL\_WGO suffer a serious stability issue, which verifies that the online buffer indeed stabilizes the learning process.
Besides, GCQL\_WG performs better than GCQL\_WGO, which indicates that the online buffer can not only stabilize the learning process but also improve the performance.
Moreover, GCQL still outperforms GCQL\_WG by a large margin on the sub-optimal dataset, which verifies the efficiency of our greedy update scheme.
In summary, greedy update schema and online buffer are both crucial for improving the sample efficiency in GCQL.

\subsection{Analysis on Hyper-parameters}\label{hyper}

As we do not use careful fine-tuning techniques (e.g., grid search) for hyper-parameters in our methods, one may wonder how the hyper-parameters affect the performance.
To this end, we conduct experiments to investigate their influence on GCQL.
Here, we use the most complicated one among three tasks, i.e., walker2d, in this experiment, and we analyze initial offline training steps $T_\text{initial}$, updating steps in each iteration $T_\text{off}$ and OORB threshold $p$ in the section.

Learning curves for analyzing $T_\text{initial}$ and $T_\text{off}$ are shown in Figure~\ref{fig:io}. We try a larger and smaller value than the default one to test its impacts. Specifically, $T_\text{initial}$ is tested with $2e5$ and $5e4$, and $T_\text{off}$ is tested with $2e4$ and $5e3$. We can see that the performance of GCQL is insensitive to the $T_\text{initial}$ and $T_\text{off}$, especially for datasets with not poor quality, e.g., datasets except for the random one. Besides, Figure~\ref{fig:p} shows curves for analyzing $p$, which have bigger impact on the performance than $T_\text{initial}$ and $T_\text{off}$.
$p$ is tested with $0.3$, $0.4$, $0.5$, $0.6$ and $0.7$, and results show that
methods with higher $p$ perform better on the medium and medium-replay datasets, while those with lower $p$ perform better on the other datasets.
Particularly, only $p=0.5$ can achieve a clear performance improvement in both the medium and random datasets.
Overall, the default $p=0.5$ is the most appropriate setting, suggesting that taking the online and offline data equally important may be the best option for GCQL in most cases.

\section{Conclusion and Future Work}

In this paper, we propose an Adaptive Policy Learning (APL) framework for offline-to-online reinforcement learning. In APL, the advantages of online and offline data are considered in an adaptive way, so that they are well-utilized for policy learning. 
Furthermore, a value-based and a policy-based algorithm are provided for implementing APL framework.
Finally, we conduct comprehensive experiments and results demonstrate that our methods can obtain best sample efficiency in the offline-to-online setting comparing with several competitive baselines.
In the future, we will continue to further improve the robustness of APL. For example, we will try to minimize the impact of offline dataset's quality on the performance.
At last, we hope this work could bridge the gap between offline and online RL.

\bibliography{aaai2023_conference} 
\end{document}

%% file: math_commands.tex
\usepackage{amsmath,amsfonts,bm}

\def\eqref#1{equation~\ref{#1}}

\def\1{\bm{1}}

\DeclareMathAlphabet{\mathsfit}{\encodingdefault}{\sfdefault}{m}{sl}
\SetMathAlphabet{\mathsfit}{bold}{\encodingdefault}{\sfdefault}{bx}{n}













%% file: aaai2023_conference.bbl
\begin{thebibliography}{35}
\providecommand{\natexlab}[1]{#1}

\bibitem[{Agarwal, Schuurmans, and Norouzi(2020)}]{agarwal2020optimistic}
Agarwal, R.; Schuurmans, D.; and Norouzi, M. 2020.
\newblock An optimistic perspective on offline reinforcement learning.
\newblock In \emph{ICML}.

\bibitem[{Brown et~al.(2020)Brown, Mann, Ryder, Subbiah, Kaplan, Dhariwal,
  Neelakantan, Shyam, Sastry, Askell et~al.}]{gpt3}
Brown, T.~B.; Mann, B.; Ryder, N.; Subbiah, M.; Kaplan, J.; Dhariwal, P.;
  Neelakantan, A.; Shyam, P.; Sastry, G.; Askell, A.; et~al. 2020.
\newblock Language models are few-shot learners.
\newblock \emph{arXiv preprint arXiv:2005.14165}.

\bibitem[{Chen et~al.(2021{\natexlab{a}})Chen, Wang, Zhou, and Ross}]{redq}
Chen, X.; Wang, C.; Zhou, Z.; and Ross, K.~W. 2021{\natexlab{a}}.
\newblock Randomized Ensembled Double Q-Learning: Learning Fast Without a
  Model.
\newblock \emph{ICLR}, abs/2101.05982.

\bibitem[{Chen et~al.(2021{\natexlab{b}})Chen, Yu, Li, Luo, Qin, Shang, and
  Ye}]{chen2021offline}
Chen, X.-H.; Yu, Y.; Li, Q.; Luo, F.-M.; Qin, Z.; Shang, W.; and Ye, J.
  2021{\natexlab{b}}.
\newblock Offline Model-based Adaptable Policy Learning.
\newblock \emph{Advances in Neural Information Processing Systems}, 34:
  8432--8443.

\bibitem[{Dann et~al.(2014)Dann, Neumann, Peters et~al.}]{opesurvey}
Dann, C.; Neumann, G.; Peters, J.; et~al. 2014.
\newblock Policy evaluation with temporal differences: A survey and comparison.
\newblock \emph{Journal of Machine Learning Research}.

\bibitem[{Deng et~al.(2009)Deng, Dong, Socher, Li, Li, and Fei-Fei}]{ImageNet}
Deng, J.; Dong, W.; Socher, R.; Li, L.-J.; Li, K.; and Fei-Fei, L. 2009.
\newblock ImageNet: A large-scale hierarchical image database.
\newblock In \emph{2009 IEEE Conference on Computer Vision and Pattern
  Recognition}, 248--255.

\bibitem[{Fu et~al.(2020)Fu, Kumar, Nachum, Tucker, and Levine}]{fu2020d4rl}
Fu, J.; Kumar, A.; Nachum, O.; Tucker, G.; and Levine, S. 2020.
\newblock D4rl: Datasets for deep data-driven reinforcement learning.
\newblock \emph{arXiv preprint arXiv:2004.07219}.

\bibitem[{Fujimoto and Gu(2021)}]{td3bc}
Fujimoto, S.; and Gu, S.~S. 2021.
\newblock A Minimalist Approach to Offline Reinforcement Learning.
\newblock arXiv:2106.06860.

\bibitem[{Fujimoto, Meger, and Precup(2019)}]{Scottbcq}
Fujimoto, S.; Meger, D.; and Precup, D. 2019.
\newblock Off-Policy Deep Reinforcement Learning without Exploration.
\newblock In \emph{ICML}.

\bibitem[{Fujimoto, van Hoof, and Meger(2018)}]{td3}
Fujimoto, S.; van Hoof, H.; and Meger, D. 2018.
\newblock Addressing Function Approximation Error in Actor-Critic Methods.
\newblock In \emph{International Conference on Machine Learning}.

\bibitem[{Ghosh et~al.(2022)Ghosh, Ajay, Agrawal, and
  Levine}]{ghosh2022offline}
Ghosh, D.; Ajay, A.; Agrawal, P.; and Levine, S. 2022.
\newblock Offline rl policies should be trained to be adaptive.
\newblock In \emph{International Conference on Machine Learning}, 7513--7530.
  PMLR.

\bibitem[{Haarnoja et~al.(2018)Haarnoja, Zhou, Abbeel, and Levine}]{sac}
Haarnoja, T.; Zhou, A.; Abbeel, P.; and Levine, S. 2018.
\newblock Soft actor-critic: Off-policy maximum entropy deep reinforcement
  learning with a stochastic actor.
\newblock \emph{arXiv preprint arXiv:1801.01290}.

\bibitem[{He and Hou(2020)}]{popo}
He, Q.; and Hou, X. 2020.
\newblock POPO: Pessimistic Offline Policy Optimization.
\newblock \emph{ArXiv}, abs/2012.13682.

\bibitem[{Jaques et~al.(2019)Jaques, Ghandeharioun, Shen, Ferguson, Lapedriza,
  Jones, Gu, and Picard}]{dialog}
Jaques, N.; Ghandeharioun, A.; Shen, J.~H.; Ferguson, C.; Lapedriza, A.; Jones,
  N.; Gu, S.; and Picard, R. 2019.
\newblock Way off-policy batch deep reinforcement learning of implicit human
  preferences in dialog.
\newblock \emph{arXiv preprint arXiv:1907.00456}.

\bibitem[{Kidambi et~al.(2020)Kidambi, Rajeswaran, Netrapalli, and
  Joachims}]{Kidambi2020MOReLM}
Kidambi, R.; Rajeswaran, A.; Netrapalli, P.; and Joachims, T. 2020.
\newblock MOReL : Model-Based Offline Reinforcement Learning.
\newblock \emph{NeurIPS}, abs/2005.05951.

\bibitem[{Kim et~al.(2013)Kim, massoud Farahmand, Pineau, and
  Precup}]{Kim2013LearningFL}
Kim, B.; massoud Farahmand, A.; Pineau, J.; and Precup, D. 2013.
\newblock Learning from Limited Demonstrations.
\newblock In \emph{NeurIPS}.

\bibitem[{Kostrikov, Nair, and Levine(2022)}]{iql}
Kostrikov, I.; Nair, A.; and Levine, S. 2022.
\newblock Offline reinforcement learning with implicit q-learning.
\newblock \emph{International Conference on Learning Representations}.

\bibitem[{Kotsiantis et~al.(2007)Kotsiantis, Zaharakis, Pintelas et~al.}]{sl}
Kotsiantis, S.~B.; Zaharakis, I.; Pintelas, P.; et~al. 2007.
\newblock Supervised machine learning: A review of classification techniques.
\newblock \emph{Emerging artificial intelligence applications in computer
  engineering}, 160(1): 3--24.

\bibitem[{Kumar et~al.(2019)Kumar, Fu, Soh, Tucker, and Levine}]{bear}
Kumar, A.; Fu, J.; Soh, M.; Tucker, G.; and Levine, S. 2019.
\newblock Stabilizing off-policy q-learning via bootstrapping error reduction.
\newblock In \emph{Advances in Neural Information Processing Systems},
  11761--11771.

\bibitem[{Kumar et~al.(2020)Kumar, Zhou, Tucker, and Levine}]{cql}
Kumar, A.; Zhou, A.; Tucker, G.; and Levine, S. 2020.
\newblock Conservative Q-Learning for Offline Reinforcement Learning.
\newblock \emph{NeurIPS}.

\bibitem[{Lee et~al.(2021)Lee, Seo, Lee, Abbeel, and
  Shin}]{lee2021offlineonline}
Lee, S.; Seo, Y.; Lee, K.; Abbeel, P.; and Shin, J. 2021.
\newblock Offline-to-Online Reinforcement Learning via Balanced Replay and
  Pessimistic Q-Ensemble.
\newblock \emph{arXiv preprint arXiv:2107.00591}.

\bibitem[{Levine et~al.(2020)Levine, Kumar, Tucker, and
  Fu}]{Levine2020OfflineRL}
Levine, S.; Kumar, A.; Tucker, G.; and Fu, J. 2020.
\newblock Offline Reinforcement Learning: Tutorial, Review, and Perspectives on
  Open Problems.
\newblock \emph{ArXiv}, abs/2005.01643.

\bibitem[{Mandel et~al.(2014)Mandel, Liu, Levine, Brunskill, and
  Popovic}]{education}
Mandel, T.; Liu, Y.-E.; Levine, S.; Brunskill, E.; and Popovic, Z. 2014.
\newblock Offline policy evaluation across representations with applications to
  educational games.
\newblock In \emph{AAMAS}, volume 1077.

\bibitem[{Matsushima et~al.(2021)Matsushima, Furuta, Matsuo, Nachum, and
  Gu}]{Deployment-efficient}
Matsushima, T.; Furuta, H.; Matsuo, Y.; Nachum, O.; and Gu, S. 2021.
\newblock Deployment-efficient reinforcement learning via model-based offline
  optimization.
\newblock \emph{ICLR}.

\bibitem[{Mnih et~al.(2015)Mnih, Kavukcuoglu, Silver, Rusu, Veness, Bellemare,
  Graves, Riedmiller, Fidjeland, Ostrovski et~al.}]{dqn}
Mnih, V.; Kavukcuoglu, K.; Silver, D.; Rusu, A.~A.; Veness, J.; Bellemare,
  M.~G.; Graves, A.; Riedmiller, M.; Fidjeland, A.~K.; Ostrovski, G.; et~al.
  2015.
\newblock Human-level control through deep reinforcement learning.
\newblock \emph{nature}, 518(7540): 529--533.

\bibitem[{Nair et~al.(2020)Nair, Dalal, Gupta, and Levine}]{awac}
Nair, A.; Dalal, M.; Gupta, A.; and Levine, S. 2020.
\newblock Accelerating Online Reinforcement Learning with Offline Datasets.
\newblock \emph{CoRR}, abs/2006.09359.

\bibitem[{Rajeswaran et~al.(2018)Rajeswaran, Kumar, Gupta, Vezzani, Schulman,
  Todorov, and Levine}]{rajeswaran2017learning}
Rajeswaran, A.; Kumar, V.; Gupta, A.; Vezzani, G.; Schulman, J.; Todorov, E.;
  and Levine, S. 2018.
\newblock Learning complex dexterous manipulation with deep reinforcement
  learning and demonstrations.
\newblock \emph{Robotics: Science and Systems}.

\bibitem[{Schulman et~al.(2015)Schulman, Levine, Abbeel, Jordan, and
  Moritz}]{trpo}
Schulman, J.; Levine, S.; Abbeel, P.; Jordan, M.; and Moritz, P. 2015.
\newblock Trust region policy optimization.
\newblock In \emph{International conference on machine learning}, 1889--1897.
  PMLR.

\bibitem[{Schulman et~al.(2017)Schulman, Wolski, Dhariwal, Radford, and
  Klimov}]{ppo}
Schulman, J.; Wolski, F.; Dhariwal, P.; Radford, A.; and Klimov, O. 2017.
\newblock Proximal policy optimization algorithms.
\newblock \emph{arXiv preprint arXiv:1707.06347}.

\bibitem[{Sutton and Barto(2011)}]{sutton2011reinforcement}
Sutton, R.~S.; and Barto, A.~G. 2011.
\newblock Reinforcement learning: An introduction.

\bibitem[{Todorov, Erez, and Tassa(2012)}]{todorov2012mujoco}
Todorov, E.; Erez, T.; and Tassa, Y. 2012.
\newblock Mujoco: A physics engine for model-based control.
\newblock In \emph{2012 IEEE/RSJ International Conference on Intelligent Robots
  and Systems}, 5026--5033. IEEE.

\bibitem[{Vecerik et~al.(2017)Vecerik, Hester, Scholz, Wang, Pietquin, Piot,
  Heess, Roth{\"o}rl, Lampe, and Riedmiller}]{vecerik2017leveraging}
Vecerik, M.; Hester, T.; Scholz, J.; Wang, F.; Pietquin, O.; Piot, B.; Heess,
  N.; Roth{\"o}rl, T.; Lampe, T.; and Riedmiller, M. 2017.
\newblock Leveraging demonstrations for deep reinforcement learning on robotics
  problems with sparse rewards.
\newblock \emph{arXiv preprint arXiv:1707.08817}.

\bibitem[{Wu, Tucker, and Nachum(2019)}]{Wu2019BehaviorRO}
Wu, Y.; Tucker, G.; and Nachum, O. 2019.
\newblock Behavior Regularized Offline Reinforcement Learning.
\newblock \emph{ArXiv}, abs/1911.11361.

\bibitem[{Yu et~al.(2020)Yu, Thomas, Yu, Ermon, Zou, Levine, Finn, and
  Ma}]{Yu2020MOPOMO}
Yu, T.; Thomas, G.; Yu, L.; Ermon, S.; Zou, J.; Levine, S.; Finn, C.; and Ma,
  T. 2020.
\newblock MOPO: Model-based Offline Policy Optimization.
\newblock \emph{NeurIPS}, abs/2005.13239.

\bibitem[{Zhu et~al.(2019)Zhu, Gupta, Rajeswaran, Levine, and
  Kumar}]{zhu2019dexterous}
Zhu, H.; Gupta, A.; Rajeswaran, A.; Levine, S.; and Kumar, V. 2019.
\newblock Dexterous manipulation with deep reinforcement learning: Efficient,
  general, and low-cost.
\newblock In \emph{2019 International Conference on Robotics and Automation
  (ICRA)}, 3651--3657. IEEE.

\end{thebibliography}
